\documentclass[letterpaper]{article} % DO NOT CHANGE THIS
\usepackage{aaai2027} % Add [submission] to enable anonymous review mode.
\nocopyright % 去掉脚注里AAAI的copyright
\usepackage[hyphens]{url} % DO NOT CHANGE THIS
\usepackage{graphicx} % DO NOT CHANGE THIS
\usepackage{natbib} % DO NOT CHANGE THIS AND DO NOT ADD OPTIONS
\usepackage{caption} % DO NOT CHANGE THIS AND DO NOT ADD OPTIONS
\usepackage{amsmath}
\usepackage{amssymb}
\usepackage{booktabs}
\usepackage{microtype}
\usepackage{algorithm}
\usepackage{algpseudocode}
\usepackage{tikz}

\newcommand{\method}{LeafSkill}
\newcommand{\recap}{ReCAP}

\newcommand{\validset}{\mathcal{A}_t}
\newcommand{\skilllib}{\mathcal{L}}

\begin{document}

\title{LeafSkill: Efficient Leaf-Level Action Grounding for Recursive LLM Agents}
% \author{Anonymous Submission}
% \affiliations{}
\author{
    Shen Liu\textsuperscript{1},
    Ze Zhou\textsuperscript{2},
    Zhenguo Xu\textsuperscript{3},
    Shaopu Wang\textsuperscript{4},
    Yike Gao\textsuperscript{4},
    Chunlei Wang\textsuperscript{4}\corresponding
}
\affiliations{
    \textsuperscript{1} North China Institute of Computer System Engineering, Beijing, China\\
    \textsuperscript{2} Beijing University of Aeronautics and Astronautics, Beijing, China\\
    \textsuperscript{3} University of Science and Technology of China, Hefei, China\\
    \textsuperscript{4} China Information Security Research Institute Co., Ltd., Beijing, China\\
    liushen24@mails.ucas.edu.cn, 22374238@buaa.edu.cn, xzg@mail.ustc.edu.cn,\\
    wangshaopu18@mails.ucas.ac.cn, yikegao@mail.ustc.edu.cn, wangchl@cecr.ac.cn
}
\maketitle

\begin{abstract}
Recursive large language model (LLM) agents often invoke a large host model after decomposition has reduced the problem to grounding a leaf task into one executable action. Our traces show that this last-mile grounding consumes 22.9--44.5\% of host-model tokens. We introduce \method{}, an external action-grounding layer that reuses local task--action experience without taking over recursive planning. \method{} separates recall from execution: semantic retrieval proposes task--action templates, deterministic binding instantiates currently admissible actions, and a lightweight controller selects a candidate or abstains. The controller cannot generate free-form commands, and abstention returns control to the host. With the same \recap{} planner and Qwen3.5-27B host, \method{} reduces host tokens on jointly solved tasks by 23.1\% on Robotouille and 41.3\% on ALFWorld, each averaged over all nine cross-run pairings of three independent repeats; an independent Robotouille set yields a 14.4\% reduction. Mean observed success is numerically higher for \method{} in both environments. A disjoint growth study expands the library from 446 to 588 skills without selector retraining or test trajectories; on tasks solved by all systems, host-token savings deepen from 18.3\% to 25.6\%. Ablations show that retrieval supplies paraphrase coverage, binding enforces executability, and learned selection controls when reuse is appropriate. Leaf-level action grounding therefore reduces expensive host inference while preserving recursive planning and recovery.
\end{abstract}

\section{Introduction}

Recursive LLM agents tackle long-horizon tasks by decomposing goals into progressively smaller task nodes~\cite{yao2023react,yao2023tot,hu2025hiagent,zhang2025recap}. Yet decomposition does not eliminate expensive host-model inference. Even after a leaf task has become explicit, the agent may still invoke the same large model merely to translate that task into one action accepted by the current environment. For example, once the planner has produced ``pick up onion1 from board1,'' the remaining decision is no longer how to solve the overall goal, but how to ground this local instruction into an executable command.

\begin{figure}[t]
    \centering
    \definecolor{leafbar}{RGB}{240,190,55}
    \definecolor{otherbar}{RGB}{100,149,198}
    \begin{tikzpicture}[x=0.068cm,y=0.56cm,font=\scriptsize]
        \draw[line width=0.7pt] (0,3.85) -- (118,3.85);
        \node[anchor=west,font=\footnotesize] at (0,3.28) {Benchmark};
        \node[anchor=center,font=\footnotesize] at (72,3.28)
            {Host-token split};
        \node[anchor=east,font=\footnotesize] at (118,3.28) {Grounding};
        \draw[black!55,line width=0.45pt] (0,2.75) -- (118,2.75);

        \node[anchor=west] at (0,2.25) {Robotouille Sync};
        \fill[otherbar] (44,1.99) rectangle (87.17,2.51);
        \fill[leafbar] (87.17,1.99) rectangle (100,2.51);
        \draw[black!35,line width=0.3pt] (44,1.99) rectangle (100,2.51);
        \node[anchor=east] at (118,2.25) {22.9\%};

        \node[anchor=west] at (0,1.35) {Robotouille Async};
        \fill[otherbar] (44,1.09) rectangle (86.72,1.61);
        \fill[leafbar] (86.72,1.09) rectangle (100,1.61);
        \draw[black!35,line width=0.3pt] (44,1.09) rectangle (100,1.61);
        \node[anchor=east] at (118,1.35) {23.7\%};

        \node[anchor=west] at (0,0.45) {ALFWorld};
        \fill[otherbar] (44,0.19) rectangle (75.06,0.71);
        \fill[leafbar] (75.06,0.19) rectangle (100,0.71);
        \draw[black!35,line width=0.3pt] (44,0.19) rectangle (100,0.71);
        \node[anchor=east] at (118,0.45) {44.5\%};

        \draw[line width=0.7pt] (0,-0.15) -- (118,-0.15);
        \fill[otherbar] (4,-0.94) rectangle (7,-0.58);
        \node[anchor=west] at (9,-0.76) {other host work};
        \fill[leafbar] (59,-0.94) rectangle (62,-0.58);
        \node[anchor=west] at (64,-0.76)
            {strict leaf/action grounding};
    \end{tikzpicture}
    \caption{Host-token composition of \recap{} trajectories used for skill-library construction in Robotouille and ALFWorld.}
    \label{fig:last-mile-footprint}
\end{figure}
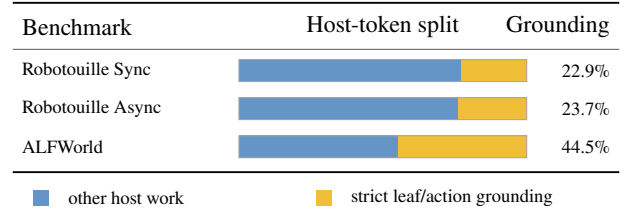

We call this operation \emph{leaf-level action grounding}: mapping an already specified task node to one currently admissible environment action, without altering decomposition, parent plans, or completion state. Although local, this operation recurs throughout a trajectory. A conservative audit of the source corpora used for library construction shows that strict leaf grounding consumes 22.9\% and 23.7\% of host-model tokens in two Robotouille splits, and 44.5\% in ALFWorld (Figure~\ref{fig:last-mile-footprint}). Leaf grounding is therefore a narrow but substantial target for reducing host inference. We study this opportunity using \recap{} as the recursive host, while leaving its planning, recovery, and completion mechanisms unchanged.

Reusing these decisions is not equivalent to exact-string caching. Semantically equivalent tasks may be expressed by different phrases, while similar phrases may encode different intents. Retrieved arguments may be absent from the current state, and the same task may ground to different actions under different observations. Effective reuse must therefore solve four distinct problems: semantic recall, state-dependent instantiation, candidate disambiguation, and safe abstention.

We introduce \method{}, an external action-grounding layer for recursive LLM agents. Its frozen library stores canonical-task/action-template pairs, each of which we call a \emph{skill}; multiple skills may share one canonical task key. Given a leaf task, semantic retrieval recalls related templates, deterministic binding instantiates them with entities from the current state and removes inadmissible actions, and a lightweight controller selects among the remaining request-local candidates or abstains. The controller cannot generate free-form commands, and abstention returns control to the host without changing agent state. \method{} thus delegates only the final choice within a finite, state-bound action set, while recursive planning, recovery, and goal completion remain entirely host-owned. This boundary distinguishes \method{} from systems that return retrieved text for host reinterpretation or route an entire prompt to another model.

With the same \recap{} planner and Qwen3.5-27B host, \method{} reduces host-model tokens on jointly solved tasks by 23.1\% on Robotouille and 41.3\% on ALFWorld, each averaged over all nine cross-run pairings of three independent repeats. The nine pairing-specific reductions span 16.8--29.7\% and 36.7--45.2\%, respectively. An independent Robotouille evaluation yields a further 14.4\% reduction, while mean observed success is numerically higher for \method{} in both environments. A disjoint growth study expands the library from 446 to 588 skills without selector retraining or test trajectories; on tasks solved by all systems, host-token savings deepen from 18.3\% to 25.6\%. Ablations further show that retrieval supplies paraphrase coverage, binding enforces executability, and learned selection controls when reuse is appropriate.

Our contributions are:
\begin{itemize}
    \item \textbf{Problem formulation and measurement.} We formulate leaf-level action grounding as a distinct delegation boundary in recursive LLM agents and show that it accounts for 22.9--44.5\% of host-model tokens in our source traces.
    \item \textbf{Constrained grounding architecture.} We develop a retrieve--bind--select layer that converts reusable task--action experience into state-bound admissible candidates, while preserving host-owned planning and recovery through explicit abstention.
    \item \textbf{Cross-environment efficiency evidence.} Across Robotouille and ALFWorld, we demonstrate substantial host-token reductions on jointly solved tasks, identify the complementary roles of retrieval, binding, and selection, and show that post-training library growth can further increase savings without selector retraining.
\end{itemize}

\section{Related Work}

\paragraph{Recursive agent backbones.}
ReAct interleaves reasoning with interaction~\cite{yao2023react}, while Tree of Thoughts searches over intermediate reasoning states~\cite{yao2023tot}. HiAgent manages long-horizon context through hierarchical working memory~\cite{hu2025hiagent}, and \recap{} represents goals, subtasks, and feedback in a recursive task tree~\cite{zhang2025recap}. These approaches primarily improve planning, context management, and recovery. \method{} is complementary: it operates only after an environment adapter exposes an eligible leaf task, without taking over decomposition or tree-state control. We instantiate this boundary with \recap{} as the recursive host.

\paragraph{Experience and skill reuse.}
Agents reuse prior experience as verbal feedback or distilled lessons~\cite{shinn2023reflexion,zhao2024expel}, executable programs or workflows~\cite{wang2024voyager,wang2025awm}, and retrieved demonstrations~\cite{kim2024rada}. RAG externalizes evidence from parametric generation~\cite{lewis2020rag}. Agent-oriented work includes ExpRAG, which retrieves trajectories into an action-generating context~\cite{ferraz2026retrievalagents}, and SkillRet, which benchmarks skill retrieval~\cite{cho2026skillret}. Recent systems further structure experience as dependency-aware skill documents or executable multi-step graphs: Graph of Skills retrieves skill bundles and graph neighbors~\cite{liu2026graphskills}, whereas GraSP compiles typed skill DAGs with node-level execution and repair~\cite{xia2026grasp}.

\method{} differs in both reuse granularity and delegated authority (Table~\ref{tab:reuse-boundaries}). It retrieves task--action templates, binds them to the current state, and permits the local controller to choose only among the resulting admissible actions or abstain. Candidate identities are request-local rather than members of a fixed global output vocabulary, so compatible skills can be added without changing the selector's output space.

\begin{table}[t]
\centering
\scriptsize
\setlength{\tabcolsep}{2.5pt}
\begin{tabular}{p{.18\columnwidth}p{.31\columnwidth}p{.41\columnwidth}}
\toprule
System & Reused unit & Delegated decision\\
\midrule
ExpRAG & Retrieved trajectories & Policy conditions action generation on retrieved experience\\
GoS & Skill documents and graph neighbors & Host interprets a dependency-aware skill bundle\\
GraSP & Typed skill DAG & Graph executor composes and repairs multiple nodes\\
\method{} & State-bound action candidates & Local selector chooses an admissible action or abstains\\
\bottomrule
\end{tabular}
\caption{Reuse granularity and delegated decisions in representative experience-reuse systems. \method{} delegates only a request-local choice over admissible actions.}
\label{tab:reuse-boundaries}
\end{table}

\paragraph{Constrained action grounding.}
Affordance-aware methods combine language-model proposals with external feasibility signals~\cite{ahn2022saycan}. SCLPlan uses PDDL-defined symbolic oversight to constrain goal and plan construction~\cite{byrd2025sclplan}, while CRANE constrains language-model generation through grammar-constrained decoding~\cite{banerjee2025crane}. \method{} targets a narrower boundary: retrieval proposes reusable templates, deterministic binding converts them into currently admissible actions, and the learned controller selects only within that finite set or abstains. It therefore constrains one leaf-level action without modifying the host planner or its decoding process.

\paragraph{Cost-aware model routing.}
Cost-aware systems select or sequence models for a complete request~\cite{chen2024frugalgpt,ong2025routellm}. \method{} does not route the recursive prompt: the host first determines the leaf task, and the local controller chooses only among state-bound actions before control returns to the same host.

\section{Method}

\subsection{Overview}

\begin{figure*}[t]
    \centering
    \includegraphics[width=\textwidth]{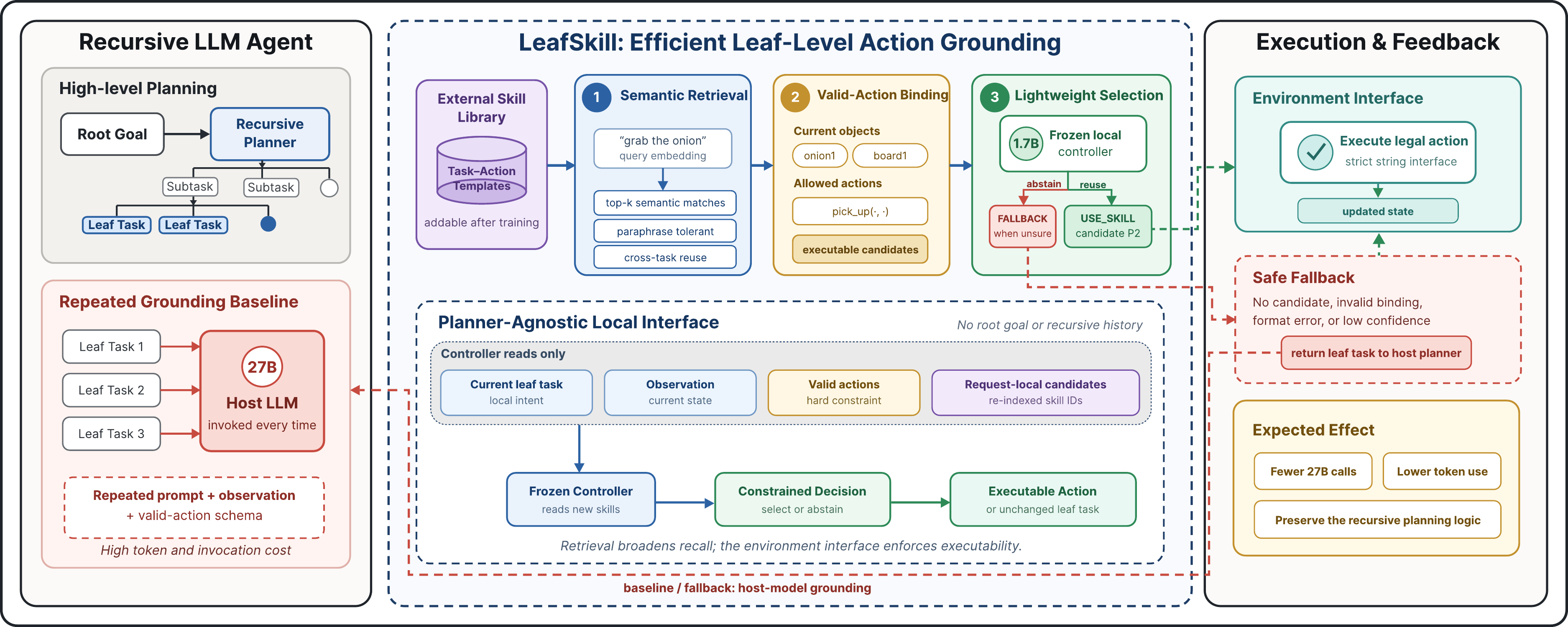}
    \caption{Overview of \method{}. At an eligible leaf task, retrieval recalls templates, binding instantiates admissible actions, and the local selector executes a candidate or abstains to the unchanged host. Source trajectories construct the frozen library and selector supervision offline.}
    \label{fig:method-overview}
\end{figure*}

\method{} has an offline construction stage and an online grounding stage. Offline, host trajectories are converted into canonical task--action templates and selector supervision. Online, \method{} is invoked only at adapter-eligible leaf tasks: it retrieves related templates, binds them to the current admissible action set, and either selects one action or abstains. Figure~\ref{fig:method-overview} summarizes both stages.

\subsection{Leaf-Level Delegation Boundary}

\noindent\fbox{\parbox{\dimexpr\columnwidth-2\fboxsep-2\fboxrule\relax}{\small
\textbf{Definition (leaf-level grounding).}
Leaf-level grounding maps an already specified task node to one action in the environment's current admissible set, without modifying task decomposition, a parent's plan, or task-completion state.
}}

Let the eligible task node, local observation, and admissible action set at time $t$ be $\tau_t$, $o_t$, and $\validset$. With host planner $P$ and external library $\skilllib$, define
\begin{equation}
\begin{aligned}
g_t &= G_\theta(\tau_t,o_t,\validset;\skilllib)
      \in \validset\cup\{\bot\},\\
\Pi_t &=
\begin{cases}
g_t, & g_t\in\validset,\\
P(\tau_t,o_t,\validset), & g_t=\bot.
\end{cases}
\end{aligned}
\label{eq:local-policy}
\end{equation}
Here $\bot$ is a state-preserving abstention: no delegated action is executed, and the same node returns to the host. The environment adapter determines \emph{where} delegation may be attempted; the selector determines \emph{whether} a bound candidate should execute.

\paragraph{Design principles.}
\emph{Narrow delegation} keeps decomposition, recovery, and completion with the host. \emph{Retrieval for recall} proposes semantically related experience but cannot authorize execution. \emph{Binding for admissibility} instantiates current entities and removes commands outside $\validset$. \emph{Selection with abstention} resolves the remaining ambiguity while preserving the host path when local evidence is insufficient.

\subsection{Retrieve--Bind--Select}

Each skill $s\in\skilllib$ is a canonical-task/action-template pair with observed surface forms and provenance. Multiple skills may share one canonical task key, preserving state-dependent one-to-many actions. For an eligible request, retrieval and binding construct
\begin{equation}
\begin{aligned}
\mathcal{R}_t
    &= \operatorname{TopK}(\tau_t,\skilllib;k,\gamma),\\
\mathcal{C}_t
    &= \{(s,a)\mid s\in\mathcal{R}_t,\;a\in\validset,\\
    &\qquad a\in\operatorname{Instantiate}(s,\tau_t,o_t)\}.
\end{aligned}
\label{eq:candidate-construction}
\end{equation}

\paragraph{Semantic retrieval.}
The retriever ranks canonical task keys against $\tau_t$ and returns the top-$k$ keys above similarity threshold $\gamma$. All templates associated with those keys proceed to binding. Retrieval is recall oriented: it may propose multiple related intents, but a retrieved template alone cannot execute.

\paragraph{State-bound binding.}
The binder instantiates typed template slots with entities from the current request and retains only concrete commands that exactly match an action in $\validset$. It merges duplicate skill--action pairs, producing the finite candidate set $\mathcal{C}_t$.

\paragraph{Local selection and abstention.}
Candidates receive request-local identifiers and are shuffled before presentation to the selector. The selector chooses one identifier or $\bot$; it cannot generate a free-form command. The selected identifier is resolved through the request-local mapping and checked against $\validset$ again before execution. The selector is not a policy over the full admissible set: retrieval and binding define its request-local support, and an empty support returns control to the host. Algorithm~\ref{alg:leafskill} gives the complete online procedure.

\begin{algorithm}[t]
\caption{Selective leaf-level grounding}
\label{alg:leafskill}
\footnotesize
\begin{algorithmic}[1]
\Require node $\tau_t$, observation $o_t$, admissible set $\validset$, library $\skilllib$
\State \textbf{if} $\neg\Call{Eligible}{\tau_t,o_t}$ \textbf{then return} $\bot$
\State $\mathcal{R}\gets\Call{TopK}{\tau_t,\skilllib,k,\gamma}$
\State $\mathcal{C}\gets\varnothing$
\ForAll{$s\in\mathcal{R},\ a\in\Call{Instantiate}{s,\tau_t,o_t}$}
    \If{$a\in\validset$}
        \State $\mathcal{C}\gets\Call{Merge}{\mathcal{C},(s,a)}$
    \EndIf
\EndFor
\State \textbf{if} $\mathcal{C}=\varnothing$ \textbf{then return} $\bot$
\State $(\widetilde{\mathcal{C}},\pi)\gets\Call{LocalizeAndShuffle}{\mathcal{C}}$
\State $j\gets\Call{Select}{\tau_t,o_t,\validset,\widetilde{\mathcal{C}}}$
\State \textbf{if} $j=\bot$ \textbf{then return} $\bot$
\State $a\gets\Call{Resolve}{j,\pi}$
\State \textbf{if} $a=\bot$ \textbf{or} $a\notin\validset$ \textbf{then return} $\bot$
\State \Return $a$
\end{algorithmic}
\end{algorithm}

\subsection{Environment Adapters and State-Bound Binding}

Eligibility is structural, not a function of retrieval similarity, library support, or controller confidence. Robotouille exposes atomic leaves directly. ALFWorld has no equivalent flag, so its adapter admits only non-root nodes on the downward execution path and excludes the root, upward-return, and revisit phases. For example, \texttt{check fridge 1} is eligible while the planner descends toward an unresolved child, but not when the node is revisited during recovery. Table~\ref{tab:adapter-example} summarizes the two adapters.

\begin{table}[t]
\centering
\scriptsize
\setlength{\tabcolsep}{2.5pt}
\begin{tabular}{p{.18\columnwidth}p{.31\columnwidth}p{.41\columnwidth}}
\toprule
Environment & Adapter boundary & Binding example\\
\midrule
Robotouille & Native atomic-leaf flag & \texttt{pick up [food] from [board]} $\rightarrow$ \texttt{pick up onion1 from board1}\\
ALFWorld & Non-root downward phase; exclude root, upward return, and revisit & \texttt{check fridge 1} $\rightarrow$ \texttt{go to fridge 1}\\
\bottomrule
\end{tabular}
\caption{Environment-specific eligibility and state-bound action instantiation.}
\label{tab:adapter-example}
\end{table}

\paragraph{Canonicalization and binding.}
Canonicalization lowercases text, joins spaced object identifiers (e.g., \texttt{fridge 1}$\rightarrow$\texttt{fridge1}), tokenizes alphanumeric fields, and replaces instance suffixes with typed placeholders (\texttt{onion1}$\rightarrow$\texttt{onion\#}). It normalizes entity indices but leaves paraphrase matching to retrieval. For each retrieved skill, the binder compares its canonical action with canonicalized strings in $\validset$ and requires every concrete object identifier mentioned by $\tau_t$ to occur in the action. It therefore enumerates current interface strings rather than asking a model to synthesize slot values.

\subsection{Selector and Deterministic Guards}

The selector receives only the task node, local observation, admissible actions, and localized candidate descriptions. It does not receive the root goal, ancestor path, remaining plan, persistent skill identifiers, or historical support statistics. All candidates surviving binding are visible, and request-local identifiers are randomized so that the selector must read candidate content rather than memorize a global label.

Empty retrieval, empty binding, explicit abstention, an unknown identifier, or a failed final admissibility check returns $\bot$. A skill associated with a failed local execution may be masked for the rest of that episode, but the mask resets between episodes and never changes the library. Table~\ref{tab:failure-containment} summarizes the deterministic return paths.

\begin{table}[t]
\centering
\scriptsize
\setlength{\tabcolsep}{2.4pt}
\begin{tabular}{p{.19\columnwidth}p{.38\columnwidth}p{.33\columnwidth}}
\toprule
Stage & Guard & Outcome\\
\midrule
Retrieval & No result above $\gamma$ & Return $\bot$\\
Binding & Missing entity, malformed command, or $a\notin\validset$ & Remove candidate; return $\bot$ if empty\\
Select/resolve & Abstention, unknown ID, or stale mapping & Return $\bot$\\
Final check & Resolved action not in $\validset$ & Return $\bot$\\
\bottomrule
\end{tabular}
\caption{Deterministic guards ensure that \method{} returns a currently admissible action or leaves execution to the host.}
\label{tab:failure-containment}
\end{table}

\subsection{Library and Supervision Construction}

A library record is extracted when the host proposes a task node and subsequently executes an environment action. Canonicalization converts the pair into a task key and an action template. This is local attribution, not a claim that the action belongs to a globally optimal trajectory.

\paragraph{Skill libraries.}
Robotouille uses 200 group-A source trees; 199 yield at least one pair, for 11,483 task--action instances. Canonicalization produces 413 task keys and 446 skills. After controller training, the growth condition unions this library with pairs from 200 group-B \recap{} trees and deduplicates canonical task--action templates. The expanded library contains 527 keys and 588 skills, adding 114 keys and 142 skills without controller updates. Group C is never used for construction, model selection, or adaptation. ALFWorld uses successful \texttt{valid\_seen} source episodes, retaining 1,091 logged instances (1,089 unique executable contexts) and producing 226 task keys and 228 skills.

\subsection{Lightweight Selector Adaptation}

SFT adapts only the Qwen3-1.7B local selector; it does not update the recursive planner or the Qwen3.5-27B host, and it does not teach free-form environment-action generation. For a deployment-shaped input
$x_i=(\tau_i,o_i,\mathcal{V}_i,\widetilde{\mathcal{C}}_i)$, the label $y_i$ is the request-local identifier whose bound action equals the action in the source trace, or $\bot$ if that action is absent. The selector is trained with
\begin{equation}
\mathcal{L}_{\mathrm{sel}}
=-\sum_i \log p_\theta(y_i\mid x_i).
\label{eq:selector-objective}
\end{equation}
Thus, adaptation supervises both \emph{which} retrieved candidate to use and \emph{when} to preserve the host fallback. Candidate identifiers are re-localized and shuffled for every example, preventing a persistent skill ID from becoming the prediction target.

\paragraph{Training data and routes.}
Training reproduces deployment-time retrieval and binding. Mixed-rank examples move the correct action away from rank one and introduce intent-conflicting distractors. Of 413 Robotouille keys, 289 may appear in the 3,177-decision training set and 124 are reserved for the post-training adoption diagnostic. ALFWorld adaptation uses the 1,089 unique executable contexts from \texttt{valid\_seen}. The default ALFWorld route starts from the Robotouille-adapted selector and then performs target-domain adaptation; the transfer-only and target-only controls omit the second and first stage, respectively. Neither Robotouille Group C nor ALFWorld \texttt{valid\_unseen} enters optimization. We train one epoch of rank-16 LoRA~\cite{hu2022lora} at each stage and freeze the resulting selector for evaluation.

\paragraph{Compatible post-training additions.}
New skills are compatible when they share the existing action grammar and typed slots, use the same retriever, bind to an exact member of $\validset$, and follow the selector's request-local description schema. Request-local identifiers remove a fixed label vocabulary but do not supply new binders or action semantics. The growth experiments therefore test read-time incorporation within one interface and template family.

\subsection{Implementation Details}

The retriever is BGE-small-en-v1.5~\cite{xiao2023bge}, and the selector is Qwen3-1.7B~\cite{yang2025qwen3}. Libraries and selector parameters are frozen during evaluation, with no online memory updates.

\section{Experiments}

We organize the evaluation around three questions. \textbf{RQ1:} How much host-model inference does \method{} save on tasks solved by both systems? \textbf{RQ2:} What distinct roles do retrieval, binding, and learned selection play? \textbf{RQ3:} Can a frozen selector exploit compatible skills added after training?

\subsection{Experimental Setup}

\paragraph{Environments and splits.}
Robotouille contains synchronous cooking tasks and asynchronous tasks with waiting and parallel dependencies~\cite{gonzalez2025robotouille}. ALFWorld provides household tasks with standard \texttt{valid\_seen} and \texttt{valid\_unseen} splits~\cite{shridhar2021alfworld}. Table~\ref{tab:data-partitions} separates construction, analysis, and final evaluation data. Robotouille repeats over Group B estimate rollout variability on the same 200 tasks and are not pooled as 600 independent instances. In the growth study only, B-group \recap{} traces are added after selector training; Group C remains evaluation only.

\begin{table}[H]
\centering
\scriptsize
\setlength{\tabcolsep}{3.0pt}
\begin{tabular}{llrp{.45\columnwidth}}
\toprule
Environment & Split & Size & Role\\
\midrule
Robotouille & A & 200 & Library and selector construction\\
             & B & 200 & Repeats, ablations, and growth source\\
             & C & 200 & Independent evaluation\\
ALFWorld & \texttt{valid\_seen} & 140 & Library and optional adaptation\\
         & \texttt{valid\_unseen} & 134 & Final evaluation\\
\bottomrule
\end{tabular}
\caption{Data partitions. Evaluation data never enter library construction or controller adaptation.}
\label{tab:data-partitions}
\end{table}

\paragraph{Systems and controller routes.}
Every system retains the \recap{} recursive planner and Qwen3.5-27B host~\cite{qwen35modelcard}. We select \recap{} as the host because its published Robotouille pass@1 exceeds CoT, ReAct, and ADaPT under the original GPT-4o protocol~\cite{zhang2025recap}; all comparisons here use our shared Qwen3.5-27B setup. Unless stated otherwise, Robotouille uses the Group-A library and Robotouille-trained selector, while ALFWorld uses the same 228-skill library with Robotouille initialization followed by \texttt{valid\_seen} adaptation. The controller analysis below separates this default from bare, transfer-only, and target-only routes.

\paragraph{Metrics.}
Success is computed over every scheduled episode, with terminated episodes counted as failures. Success differences use two-sided exact McNemar tests on all task pairs. Host efficiency is computed on tasks solved by both systems in a run pairing because failed long-horizon trajectories have heterogeneous stopping behavior, so their token counts conflate grounding cost with failure timing. We report reductions in Qwen3.5-27B prompt-plus-completion tokens and calls, termed \emph{paired-success host cost}; reporting full-set success alongside this metric prevents apparent savings from arising mechanically from earlier or more frequent failures. When both systems have three repeats, the reported reduction is the arithmetic mean of the nine cross-run pairings; tasks are intersected and the cost ratio is computed separately within each cell before averaging. A single-run ablation is paired with all three \recap{} repeats and reported as the mean of those three reductions. Group-C cost intervals use task-level bootstrap 95\% CIs.

\paragraph{Controls and training.}
End-to-end methods share task identifiers, model services, context limits, decoding settings, and termination rules. Retrieval uses $k=5$ and similarity threshold $\gamma=0.70$ throughout, with no test-set hyperparameter search. ALFWorld uses ten round-robin shards and a 49-action limit; every run records all 134 unseen games. Controller tokens are reported separately because the 1.7B and 27B tokenizers differ. Only LoRA parameters of the 1.7B selector are updated; the host, planner, retriever, and binder remain frozen. Each main-system SFT stage uses one epoch of rank-16 LoRA, batch size 1, gradient accumulation 16, learning rate $2\times10^{-5}$, and maximum length 1,536 on one A800 80GB GPU; the context-augmented input-scope diagnostic uses maximum length 1,792.

\begin{figure*}[t]
\centering
\includegraphics[width=.98\textwidth]{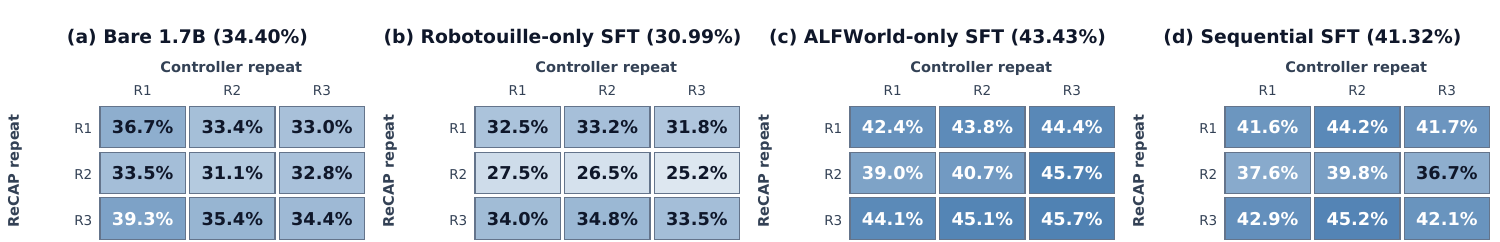}
\caption{ALFWorld cross-run paired-success host-token reduction for four controller routes. Each $3{\times}3$ panel crosses \recap{} repeats (rows) with controller repeats (columns), and its title gives the nine-cell mean $\bar{\Delta}_9$. Jointly solved counts span 106--109, 112--116, 107--111, and 114--117 in panels (a)--(d).}
\label{fig:alf-controller-matrices}
\end{figure*}

\subsection{End-to-End Host Efficiency (RQ1)}

\begin{figure}[t]
\centering
\includegraphics[width=\columnwidth]{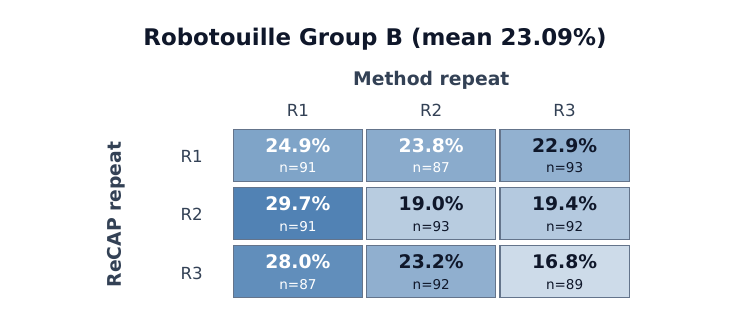}
\caption{Robotouille cross-run paired-success host-token reduction. Rows index \recap{} repeats and columns index \method{} repeats; each cell reports its pairing-specific reduction and jointly solved count $n$. The title reports the arithmetic mean over all nine cross-run pairings, $\bar{\Delta}_9$.}
\label{fig:cross-run-matrix}
\end{figure}

\begin{table}[t]
\centering
\scriptsize
\setlength{\tabcolsep}{5.0pt}
\renewcommand{\arraystretch}{0.98}
\begin{tabular}{@{}lrrr@{}}
\toprule
Metric & $n$ & Reduction & 95\% CI\\
\midrule
Tokens & 106 & \textbf{14.4\%} & $[3.0,24.5]$\\
Calls & 106 & \textbf{14.1\%} & $[3.6,23.3]$\\
\bottomrule
\end{tabular}
\caption{Independent Robotouille-C paired-success host-cost reduction.}
\label{tab:independent-cost}
\end{table}

\begin{table}[t]
\centering
\scriptsize
\setlength{\tabcolsep}{2.2pt}
\renewcommand{\arraystretch}{0.94}
\begin{tabular}{@{}llrrrr@{}}
\toprule
Evaluation & Run & \recap{} & \method{} & $p$ & $\Delta$\\
\midrule
Robotouille B & R1 & 55.5 & \textbf{61.5} & .126 & +6.0\\
 & R2 & 56.5 & \textbf{65.5} & .025 & +9.0\\
 & R3 & 53.0 & \textbf{59.5} & .079 & +6.5\\
 & Mean & -- & -- & -- & \textbf{+7.2}\\
Robotouille C & R1 & 60.5 & \textbf{68.5} & \textbf{.0259} & \textbf{+8.0}\\
ALFWorld unseen & R1 & 90.3 & \textbf{90.3} & 1.00 & 0.0\\
 & R2 & 89.6 & \textbf{92.5} & .344 & +2.9\\
 & R3 & 89.6 & \textbf{92.5} & .424 & +2.9\\
 & Mean & -- & -- & -- & \textbf{+2.0}\\
\bottomrule
\end{tabular}
\caption{End-to-end success on all scheduled tasks. $\Delta$ is the percentage-point difference; $p$ is the two-sided exact McNemar result for each run.}
\label{tab:end-to-end-success}
\end{table}

\paragraph{Robotouille.}
Across all nine cross-run pairings of the three Group-B repeats, \method{} reduces host tokens by 23.09\% on average (range 16.85--29.68\%, $n=87$--93 per pairing) and host calls by 22.54\% on average (Figure~\ref{fig:cross-run-matrix}). On the independently seeded Group C, the frozen Group-A library reduces tokens by 14.4\% (95\% CI $[3.0,24.5]$) and calls by 14.1\% (95\% CI $[3.6,23.3]$; Table~\ref{tab:independent-cost}). Observed success is 6.0--9.0 points higher on B and 8.0 points higher on C (137/200 versus 121/200, $p=.0259$; Table~\ref{tab:end-to-end-success}).

\paragraph{ALFWorld.}
Across the nine \texttt{valid\_unseen} baseline--method pairings, the default sequential selector reduces host tokens by 41.32\% on average (range 36.73--45.21\%, $n=114$--117) and host calls by 41.75\% on average. Over all scheduled games, mean success is 91.8\% for \method{} and 89.8\% for \recap{} (Table~\ref{tab:end-to-end-success}).

\begin{table}[t]
\centering
\scriptsize
\setlength{\tabcolsep}{3.0pt}
\renewcommand{\arraystretch}{0.96}
\begin{tabular}{@{}lrrrr@{}}
\toprule
Evaluation & Ep. & Calls & Ctrl. tok. & Tok./ep.\\
\midrule
Robotouille B R1 & 200 & 1,485 & -- & --\\
Robotouille B R2--R3 & 400 & 3,849 & 3.98M & 9,952\\
Robotouille C & 200 & 1,355 & 1.38M & 6,900\\
ALFWorld unseen R1--R3 & 402 & 4,636 & 3.31M & 8,236\\
\bottomrule
\end{tabular}
\caption{Measured Qwen3-1.7B controller overhead over all scheduled episodes. Tokens include prompts and completions; Robotouille-B R1 predates token instrumentation.}
\label{tab:controller-overhead}
\end{table}

\paragraph{Controller overhead.}
Table~\ref{tab:controller-overhead} reports auxiliary inference over all scheduled episodes. Instrumented runs use 6,900--9,952 controller tokens per episode (714--1,034 per call). Because the controller and host differ in tokenizer and scale, we report them separately rather than sum raw tokens.

\paragraph{Cross-run robustness.}
Crossing every \recap{} repeat with every method repeat yields positive host-token reductions in all nine Robotouille comparisons and, for every controller route, all nine ALFWorld comparisons (Figures~\ref{fig:alf-controller-matrices} and~\ref{fig:cross-run-matrix}). Reporting $\bar{\Delta}_9$ uses every observed repeat pairing and avoids privileging an arbitrary same-index alignment.

\subsection{Component Contributions (RQ2)}

\begin{table}[t]
\centering
\scriptsize
\setlength{\tabcolsep}{2.0pt}
\renewcommand{\arraystretch}{0.96}
\begin{tabular}{@{}llrr@{}}
\toprule
Reuse path & R/B/S & Success ($p_2$) & Tok. red.\\
\midrule
\recap{} R2 & --/--/Host & 113/200 (--) & --\\
RawPair-Exact & Ex./Raw/Dir. & 121/200 (.3581) & 0.20\\
RawPair-BGE & Sem./Raw/Dir. & 127/200 (.0541) & 8.08\\
Canonical-Exact & Ex./St./SFT & \textbf{134/200 (.0075)} & \textbf{29.60}\\
BGE Top-1 & Sem./St./Dir. & 124/200 (.2000) & 13.08\\
BGE Top-5 + Base & Sem./St./Base & 113/200 (1.000) & 27.01\\
\bottomrule
\end{tabular}
\caption{End-to-end ablations on Robotouille B. Token reductions average three comparisons with \recap{} R1--R3 ($n=83$--97); success and $p_2$ compare with R2. R/B/S denotes retrieval/binding/selector.}
\label{tab:component-ablation}
\end{table}

\begin{table}[t]
\centering
\scriptsize
\setlength{\tabcolsep}{1.5pt}
\renewcommand{\arraystretch}{1.00}
\begin{tabular}{@{}p{.30\columnwidth}rp{.25\columnwidth}r@{}}
\toprule
Diagnostic & $n$ & Comparison & Result (\%)\\
\midrule
Lexical rewrite coverage & 200 & Exact / Top-1 / Top-5 & 0.0 / 79.5 / 96.0\\
Syntactic rewrite coverage & 200 & Exact / Top-1 / Top-5 & 0.0 / 70.5 / 97.0\\
Mixed-rank selection & 193 & Base / SFT & 81.9 / 93.3\\
All-negative fallback & 200 & SFT & 7.5\\
Input-scope mixed-rank & 193 & Local / + planner & 93.3 / 90.7\\
\bottomrule
\end{tabular}
\caption{Controlled diagnostics: binding coverage (top two rows) and controller accuracy (bottom three).}
\label{tab:component-diagnostics}
\end{table}

\paragraph{Controlled diagnostic protocol.}
Lexical rewrites change only the action predicate (e.g., \texttt{Place}$\rightarrow$\texttt{Set}); syntactic rewrites also reorder the command. Coverage requires the bound set to retain the source action. Mixed-rank instances add up to four admissible but intent-inconsistent distractors and place the source action last. All-negative instances bypass object-coverage binding and contain only admissible but leaf-inconsistent candidates; the correct output is \texttt{FALLBACK\_RECAP} ($\bot$). This is an OOD stress test, not an online condition. The input-scope diagnostic compares two separately trained selectors constructed from the same 3,177 examples and canonical-task split. Both are evaluated on the same 193 mixed-rank cases. The context-augmented selector additionally receives the true recursion depth, root goal, parent task, ancestor chain, and remaining siblings. Table~\ref{tab:component-diagnostics} reports all sample sizes.

\paragraph{Retrieval and binding.}
Exact reuse of raw task--action pairs changes paired host tokens by only 0.20\% on average. Semantic retrieval over the same raw pairs yields an 8.08\% reduction, while state-bound templates with direct Top-1 execution raise the reduction to 13.08\% (Table~\ref{tab:component-ablation}). Canonical-Exact with the SFT selector reaches 67.0\% success and 29.60\% mean token reduction on the original benchmark wording. Its coverage falls from 99.5\% on original reference actions to 0\% under both controlled rewrites; semantic Top-5 retains 96.0\% lexical and 97.0\% syntactic coverage (Table~\ref{tab:component-diagnostics}). Semantic retrieval therefore supplies the paraphrase coverage used by the default system.

\paragraph{Selection and abstention.}
On the 100 tasks solved by both BGE Top-1 and \method{} R2, Top-5 selection provides a further 14.9\% host-token reduction. SFT raises mixed-rank accuracy from 81.9\% to 93.3\%. End to end, Base reaches 113/200 success and 27.01\% mean token reduction; SFT averages 62.17\% success and 23.09\% reduction, and is 9.0 points more successful in the direct R2 comparison ($p=.0198$). Thus, the selector calibrates when and which bound candidate to execute. In the all-negative OOD diagnostic, SFT returns \texttt{FALLBACK\_RECAP} in only 7.5\% of cases. Because binding is bypassed, this measures controller-only fallback rather than online abstention and motivates deterministic binding before selection. Adding true planner context yields 90.7\% mixed-rank accuracy versus 93.3\% with the local interface.

\subsection{Post-Training Library Growth (RQ3)}

\begin{table}[t]
\centering
\scriptsize
\setlength{\tabcolsep}{3.0pt}
\begin{tabular}{lrrrrr}
\toprule
Library & Keys & Skills & Success & Tok. red. & Call red.\\
\midrule
A & 413 & 446 & \textbf{137/200} & 18.35\% & 17.70\%\\
A+B & 527 & 588 & 131/200 & \textbf{25.64\%} & \textbf{24.22\%}\\
\bottomrule
\end{tabular}
\caption{Frozen-selector growth on Robotouille C. Cost columns use the same 88 tasks solved by \recap{} and both library variants.}
\label{tab:library-growth}
\end{table}

After selector training, adding Group-B traces expands the library from 413 keys/446 skills to 527/588 without using Group C or updating controller parameters. On the fixed set of 88 tasks solved by all three systems, the expanded library increases the host-token reduction over \recap{} from 18.35\% to 25.64\% and the call reduction from 17.70\% to 24.22\% (Table~\ref{tab:library-growth}). Comparing the two libraries directly on their 105 common successes, A+B reduces host tokens by 11.41\% (95\% CI $[2.29,19.49]$) and calls by 10.36\% relative to A. Group-C success is 131/200 with A+B and 137/200 with A ($p=.5118$). In the controlled selector diagnostic, inserting 124 held-out compatible keys yields 97.3\% adoption and 98.9\% retention. Together, these results show that a frozen selector can use compatible additions at read time.

\subsection{Controller Transfer and Adaptation}

\begin{table}[t]
\centering
\scriptsize
\setlength{\tabcolsep}{3.0pt}
\begin{tabular}{llrr}
\toprule
Controller & Training route & Succ. (\%) & Use (\%)\\
\midrule
\recap{} & no controller & 89.80 & --\\
Bare 1.7B & none & 84.33 & 79.41\\
Robot.-only & Robot. SFT & 91.29 & 70.99\\
ALF-only & ALF seen SFT & 86.07 & 98.02\\
Sequential & Robot. $\rightarrow$ ALF seen & \textbf{91.79} & \textbf{98.81}\\
\bottomrule
\end{tabular}
\caption{ALFWorld success and skill use, averaged over three \texttt{valid\_unseen} runs. All routes use the same 228-skill ALFWorld library; cross-run token results appear in Figure~\ref{fig:alf-controller-matrices}.}
\label{tab:alf-controller}
\end{table}

Figure~\ref{fig:alf-controller-matrices} compares all four routes. Robotouille-only SFT transfers with 91.29\% mean success; ALFWorld-only SFT maximizes use (98.02\%) and token reduction (43.43\%). Sequential SFT achieves the highest success (91.79\%) with 98.81\% use and 41.32\% token reduction, and is therefore the default.

\section{Discussion and Conclusion}

\paragraph{Efficiency and growth.}
\method{} replaces repeated 27B leaf-grounding calls with a finite local choice while leaving planning and recovery to the host. Retrieval supplies paraphrase coverage, binding enforces admissibility, and selection resolves candidates. Request-local identifiers also support compatible additions: expanding the library from 446 to 588 skills deepens fixed-set host-token savings from 18.3\% to 25.6\% without retraining.

\paragraph{Scope.}
\method{} assumes recurring parameterizable leaves, a discrete admissible-action interface, and an environment adapter. New action grammars require corresponding adapters and binders, while open-ended or continuous actions may need richer executable-action interfaces. The present evidence covers one planner, one host, two environments, and one selector family.

Overall, \method{} shows that recurring leaf grounding can be separated from recursive reasoning without replacing the host planner. Across Robotouille and ALFWorld, its retrieve--bind--select layer reduces Qwen3.5-27B host tokens by 14.4--41.3\%, while compatible library growth increases savings without selector retraining. These results support state-bound action reuse as a practical complement to recursive LLM planning rather than a replacement for it.

\bibliography{references/leafskill_refs}

\end{document}